%% file: jmlr-sample.tex
 \pdfoutput=1
 \documentclass[pmlr,twocolumn]{jmlr} 

\usepackage{booktabs}
\usepackage[load-configurations=version-1]{siunitx} 

\theorembodyfont{\upshape}
\theoremheaderfont{\scshape}
\theorempostheader{:}
\theoremsep{\newline}

\jmlrvolume{ML4H Extended Abstract Arxiv Index}
\jmlryear{2020}
\jmlrpublished{}

\usepackage{amsmath}
\usepackage{amssymb}
\usepackage{floatrow}
\usepackage{paralist}
\usepackage{multirow}
\usepackage{booktabs}
\usepackage{algorithm}
\usepackage{algorithmic}

\usepackage{enumitem}

\newcommand{\Semigran}{{\it{Semigran}}}
\newcommand{\ASSEMENTEST}{{\it{COVID-Assessment}}}
\newcommand{\EXPERTTEST}{{\it{Sim-from-Expert-System}}}

\newcommand{\Base}{{\small {\it{Ours-BASE}}}}
\newcommand{\BaseCOVID}{{\small {\it{Ours-BASE-COVID}}}}
\newcommand{\BaseCOVIDFULL}{{\small {\it{Ours-BASE-COVID-FULL}}}}

\title{COVID-19 in Differential Diagnosis\\ of Online Symptom Assessments}

  \author{%
   \Name{Anitha Kannan} \Email{anitha@curai.com}\\
   \Name{Richard Chen} \Email{ricky@curai.com}\\
   \Name{Vignesh Venkataraman} \Email{viggy@curai.com}\\
   \Name{Geoffrey Chen} \Email{geoff@curai.com}\\
   \Name{Xavier Amatriain} \Email{xavier@curai.com}\\
   \addr Curai
  }

\begin{document}

\maketitle

\begin{abstract}
The COVID-19 pandemic has magnified an already existing trend of people looking for answers to their healthcare concerns online. Symptom checkers are a common approach to provide an online assessment, but the use of rule-based systems introduces important limitations. This leads to COVID-19 symptom checkers not being able to provide a differential diagnosis. In this paper we present an approach that combines the strengths of traditional AI expert systems with novel deep learning models to leverage prior knowledge as well as any amount of existing data. We combine synthetic data generated from an expert system with real-world data coming from an online COVID-19 symptom checker to train a COVID-19 aware differential diagnosis deep learning model and show that our approach is able to accurately model new incoming data about COVID-19 while still preserving accuracy on other diagnosis. 

\end{abstract}

\input{sections/intro}

\input{sections/dataset}

\input{sections/model}


\input{sections/semigran_results}

\input{sections/exp}

\vspace{-.2in}

\section{Conclusions}
In this paper we have presented a novel approach to quickly enhance a diagnosis model that is effective even when a new previously unknown condition appears and compromises prior medical knowledge. Our approach combines the strengths of two very different AI formulations: traditional expert systems, with state-of-the-art deep learning models. We leverage expert systems as a way to input prior knowledge into the learned model as synthetic data, and use deep learning to learn a generalizable model on the combination of old and new data. Our model is able to capture the nuances of a new condition like COVID-19 without losing the pre-existing medical knowledge accumulated in the expert system. Our paper also demonstrates the efficiency of the approach even in situations where there is little data available to model new diseases.



\bibliography{jmlr-sample}
\appendix
\input{sections/appendix}





\end{document}

%% file: sections/intro.tex
\section{Introduction}
AI has been connected to medicine since almost its inception when expert systems designed by doctors were introduced as medical decision support tools. Modern online COVID-19 symptom checkers, which share the same underlying rule-based approach, are not able to provide a holistic diagnostic of the patient. In other words, they cannot tell the patient that \emph{e.g.} while they are unlikely to have COVID-19 they should instead worry about strep throat. This highlights one of the main shortcoming of expert systems: they are hard to scale and lack flexibility since adding a new condition requires re-tuning the system mostly manually. A very different approach to building diagnosis models is to use data to machine learn a model (see \cite{shickel_deep_2017} for example). Large data repositories can be mined to automatically learn fine-grained diagnosis models that can be easily extended and updated as new data becomes available. On the flip side, these models will only be as good as the data on which they are trained. This becomes particularly limiting in a situation like the current one in which data is hardly available and of limited quality.

In this paper we address the question of whether we can quickly learn a generalizable diagnosis model when a new condition like COVID-19 appears. We present a machine learning approach to quickly enhance an existing AI diagnosis model to incorporate a novel disease like COVID-19. We show that the resulting model is accurate in including COVID-19 in the differential diagnosis, without losing accuracy for other diseases. Finally, we also show that the approach is easily extensible as new evidence about the new diseases are surfaced.

%% file: sections/dataset.tex
\vspace{-.2in}
\section{Datasets used in this study}

\noindent{\bf Dataset from COVID-19 assessment tool (\ASSEMENTEST) :} We have access to a publicly deployed virtual diagnostic assessment tool (Appendix~\ref{apd:simulator}, Fig.~\ref{fig:flow}). This tool guides users through a comprehensive set of clinical questions to determine the likelihood of COVID-19 infection and associated complications from the disease. The users on this system are also optionally connected to a practitioner for further evaluation of their risk. Questions in the assessment is based on guidance provided by the United States Centers for Disease Control and Prevention (CDC), and hence elicits information regarding clinical factors including demographic information, symptoms, COVID-19 exposure risks and medical history of the patient. For example, a younger user with a weakened immune system with a recent exposure to the virus would likely have milder or no symptoms as compared to an older user with lung disease with exposure seven days prior to undergoing the assessment.  

We used the derived dataset from this tool to build a dataset of COVID-19 clinical cases. In particular, the assessment flow that resulted in medium or high risk are considered as positive examples of COVID-19 (Appendix~\ref{apd:simulator}, Fig.~\ref{fig:covidsim} for example cases). We used two variants of data gathered by the assessment. In one variant, we restricted to only findings (symptoms) that are also part of the expert system, and in another variant, we used all the findings (findings unique to COVID-19 in Appendix~\ref{apd:simulator} tbl.~\ref{tbl:covidFindings}). In total, we have 100 clinical cases from this dataset of which we use 70:30 split for training and evaluation.

\noindent{\bf Dataset from expert system (\EXPERTTEST) :} We use an extended version of the QMR knowledge base~\cite{miller1990quick} that consists of 830 diseases and 2052 findings (covering symptoms, signs, and demographic variables), and their relationships. We sample cases on a per-disease basis  (c.f.\citep{qmr-simulated-cases, ravuri18}). Unlike in prior works, we label each case as a distribution over the possible diagnosis by using the inference algorithm of the expert system. In the end, we simulate 65,000 clinical cases corresponding to 437 diseases covering 1418 findings, with each disease having at least 50 clinical cases.  Details of the algorithm and examples of clinical cases are provided in Appendix~\ref{apd:expert}.
\\

\noindent{\bf \Semigran}:  This a publicly available dataset ~\cite{Semigran} on which over fifty online symptom checkers were evaluated.  The dataset consists of 45 standardized patient clinical vignettes, corresponding to 39 unique diseases. We used the simplified inputs provided along with the clinical vignettes, as previously used in other studies \cite{razzaki18,kannan20}. This dataset has been studied in \cite{Fraser18}  where twenty medical experts studied each case and came to consensus, results of which we report in tbl.~\ref{tbl:semigran}.

For training the model, we use \ASSEMENTEST ~and \EXPERTTEST. Combining these two sources allows the consideration of COVID-19 in a differential diagnosis whenever appropriate while also modeling competing hypothesis. For evaluation, we use \Semigran~ and 30\% of data from \ASSEMENTEST.

%% file: sections/model.tex
\vspace{-.2in}
\section{Approach }
\label{sec:model}
{\noindent \bf Notation}: Let  $\mathcal{F} = \{f_1, \cdots , f_K\}$  be the universe of findings/symptoms ({\it e.g} `eye pain', `nausea') that can be elicited from the patients.  Let $\mathcal{Y} =  \{1, ..., L\}$ be the universe of diagnoses ({\it e.g.} `COVID-19', `arthritis' and `common cold') A clinical case is set of findings  $\mathbf{x} = \mathbf{x}_{pos} \cup \mathbf{x}_{neg}$ and $\mathbf{p}(y|\mathbf{x})$.  $\mathbf{x}_{pos} \in \mathcal{F}$ and $\mathbf{x}_{neg} \in \mathcal{F}$ are findings that are  present and  explicitly absent, respectively, while $\mathbf{p}(y|\mathbf{x})$  captures the uncertainty in diagnosis (differential diagnosis).


{\noindent \bf Loss function}: We use Kullback-Leibler divergence  from $\textbf{g}(\mathbf{x})$ to $\mathbf{p}(y|\mathbf{x})$ 
\begin{equation}
D_{\text{KL}}(\mathbf{p}  \parallel \textbf{g})=\sum_{y\in {\mathcal {Y}}}\mathbf{p}(y|\mathbf{x}) \log (\frac {\mathbf{p}(y|\mathbf{x}) }{\textbf{g}(\mathbf{x})[y]})
 \label{eqn:loss}
\end{equation}
This optimizes to capture all disease labels $y$ for which $\mathbf{p}(y|\mathbf{x})> 0$, potentially at the expense of some erroneous diagnosis. However, this is the correct thing to do in differential diagnosis when there is only partial information - to err on the side of over prediction than as failure to consider any potential disease.   When the ground truth label is a single disease, KL divergence is same as the cross entropy criterion. 


{\noindent \bf Model architecture}: The model for $\textbf{g}(\mathbf{x})$ has separate input streams for the demographic and non-demographic input findings.

Demographic variables, such as the gender and age, impose an implicit prior over diseases that are \emph{impossible}, and serves as bottle neck for diagnoses based only on the non-demographic variables. As examples, its almost impossible for an infant to be diagnosed with `dementia', or a biologically male patient with `pregnancy'. When a biologically male patient have symptoms of nausea and vomiting, these priors guide the model (especially in early iterations) to correctly place close to zero probability mass over women-related health issues. We use dense trainable $L$-dimensional embeddings initialized with uniform prior over all plausible diseases.

Non-demographic findings are separately modeled for their presence and absence state. Each finding-state is a 1024 embedding space followed by dropout for regularization. Then, the embeddings for all the observed findings are average pooled and projected to a fully connected layer. After a log soft-max transformation, this is additively combined with log softmax transformation of demographic variable representation. 

The embedding vectors for the non-demographic findings are initialized randomly in the range [-.05,.05] and dropout of .7 to regularize the model. Models are trained with minibatches of size 512 using ADAM with initial learning rate of 0.01, and trained for fixed number of fifteen epochs. Models implemented using PyTorch is trained on single NVIDIA Tesla K80 GPU.

%% file: sections/semigran_results.tex
\begin{table*}
  \begin{tabular}{cccl}
    \toprule
    Approach & top-1 & top-3 & top-5  \\
    \midrule
    Human-Doctors \cite{Fraser18}  & 72.1\% & 84.3 \%  & - \\
    AI expert system  & 66\% & 75\% & 86\% \\
    \cite{razzaki18}   &  46.6\%  & 64.67\% & - \\
    \cite{kannan20} &50.67\% (1.86) &	75.11\% (1.86) &	82.22\% (1.56)  \\
     \Base & 67.6\% (.023) &  85.8\% (.025) & 92.9\% (.009) \\
     \BaseCOVID & 61.8\% (.029)& 84.4\% (.027)& 93.3\% (.000)  \\
     \BaseCOVIDFULL & 65.5\% (.012)& 84.4\% (.041)& 93.3\% (.000)  \\
  \bottomrule
\end{tabular}
  \caption{Comparison on \Semigran~ dataset. Standard deviation (in brackets) computed on five models trained with random initialization. \cite{razzaki18} considered only 30 clinical cases and we extrapolated assuming remaining 15 cases were wrongly diagnosed. For qualitative comparisons, please see Appendix~\ref{apd:qual} Tbl.~\ref{fig:semigranQual} }  
\label{tbl:semigran}
\end{table*}

%% file: sections/exp.tex
\section{Experiments}
\noindent\textbf{Metrics:} We report top-k accuracy \emph{a.k.a} recall@k (k $\in \{1, 3,5\}$). When evaluating model performance on cases from COVID-19 assessment data, we report accuracy of predicting COVID-19 within top-k.



\noindent\textbf{Model variants:} \Base~is trained with \EXPERTTEST~to re-establish that expert systems can be modeled as data prior through simulation. \BaseCOVID~uses \EXPERTTEST~and training set of \ASSEMENTEST, while ensuring that the universe of findings $\mathcal{F}$ is same as \Base.  \BaseCOVIDFULL~ is  same as \BaseCOVID ~ but all symptoms are used.

\subsection{Results}
{\noindent \bf Approach efficacy:} Table~\ref{tbl:semigran} compares our model to existing published results on \Semigran~ dataset. \Base ~performs best across all models, closing the gap with the AI expert system that was used to simulate the datase, while re-establishing that expert systems as a data prior continues to hold in new settings, with different datasets and machine learning model.  Adding the extra disease label (COVID-19) does not deteriorate the performance as evidenced by \BaseCOVID. We also analyzed test cases with (22 cases) and without (23 cases) overlapping symptoms with COVID-19 and found that performance drop in comparison to \Base~is only caused by cases with overlapping symptoms. \BaseCOVIDFULL~with additional COVID-19 findings do not change the prediction accuracy as much as \BaseCOVID~as it can use these additional findings specific to COVID-19. 
\begin{table}
  \begin{tabular}{c|c|c}
    \toprule
    $\textrm{top-k}$ & \it{Ours-BASE-} &\it{Ours-BASE-} \\
    &{\it COVID} & {\it COVID-FULL}  \\
    \midrule
1	&	40\%	&	87\%	\\
3	&	60\%	&	100\%	\\
5	&	73\%   &	100\%	\\
  \bottomrule
\end{tabular}
 \caption{Comparison on COVID assessment data. \Base~ and AI expert system will have {\bf 0}\%. Qualitative examples in Appendix~\ref{apd:qual} Tbl.~\ref{tbl:covidExamples}} 
\label{tbl:covid-top-k}
\end{table}

{\noindent \bf COVID-19 in differential diagnosis} Table.~\ref{tbl:covid-top-k} compares the performance of \BaseCOVID~ and \BaseCOVIDFULL~ on 
 \ASSEMENTEST~ tsst dataset. By including COVID-19 in 73\% of the cases, \BaseCOVID~ establishes that it can model COVID-19 in the light of other diseases. On examining cases where COVID-19 was not part of its prediction, we found these to mainly correspond to those cases where input findings do not sufficiently overlap with the modeled findings in \BaseCOVID, which is overcome by \BaseCOVIDFULL~ that uses all the findings.

%% file: sections/appendix.tex
\section{Qualitative results}
\label{apd:qual}
Tbl.~\ref{tbl:covidExamples} provides qualitative examples comparing the three models on the data from COVID-19 assessments. These are cases with low to moderate risk of contracting COVID-19. \Base ~and \BaseCOVID~considers only the findings in column 1 as inputs, while \BaseCOVIDFULL~use the union of the first two columns as input. In the final example, as observed in \Semigran~cases, \BaseCOVID~ differential diagnosis is consistent with \Base. However, once the model gets an added input of the patient being a healthcare worker, \BaseCOVIDFULL~ includes COVID-19 in one of its top 5 position. 

 As an example, consider first example in tbl.~\ref{fig:semigranQual}. \BaseCOVID~includes COVID-19 in the differential because of the overlapping findings between viral respiratory infections and COVID-19. In contrast, in second example, with symptoms such as neck stiffness, severe headache and photophobia, the model continues to maintain its prediction to be more close to the differential diagnosis of \Base. 
\input{sections/qual1}

\input{sections/qual2}


\section{\ASSEMENTEST}
\begin{figure}
\minipage{0.25\textwidth}
  \includegraphics[width=\linewidth]{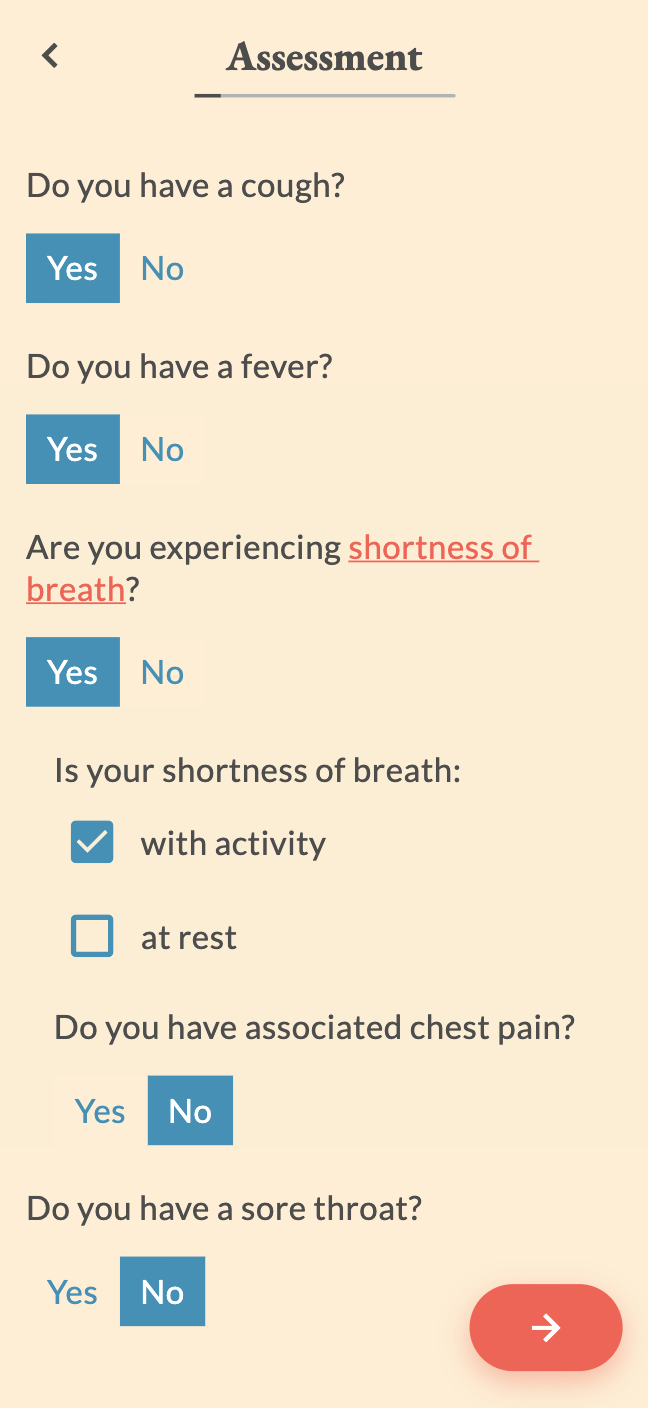}
\endminipage\hfill
\minipage{0.25\textwidth}
  \includegraphics[width=\linewidth]{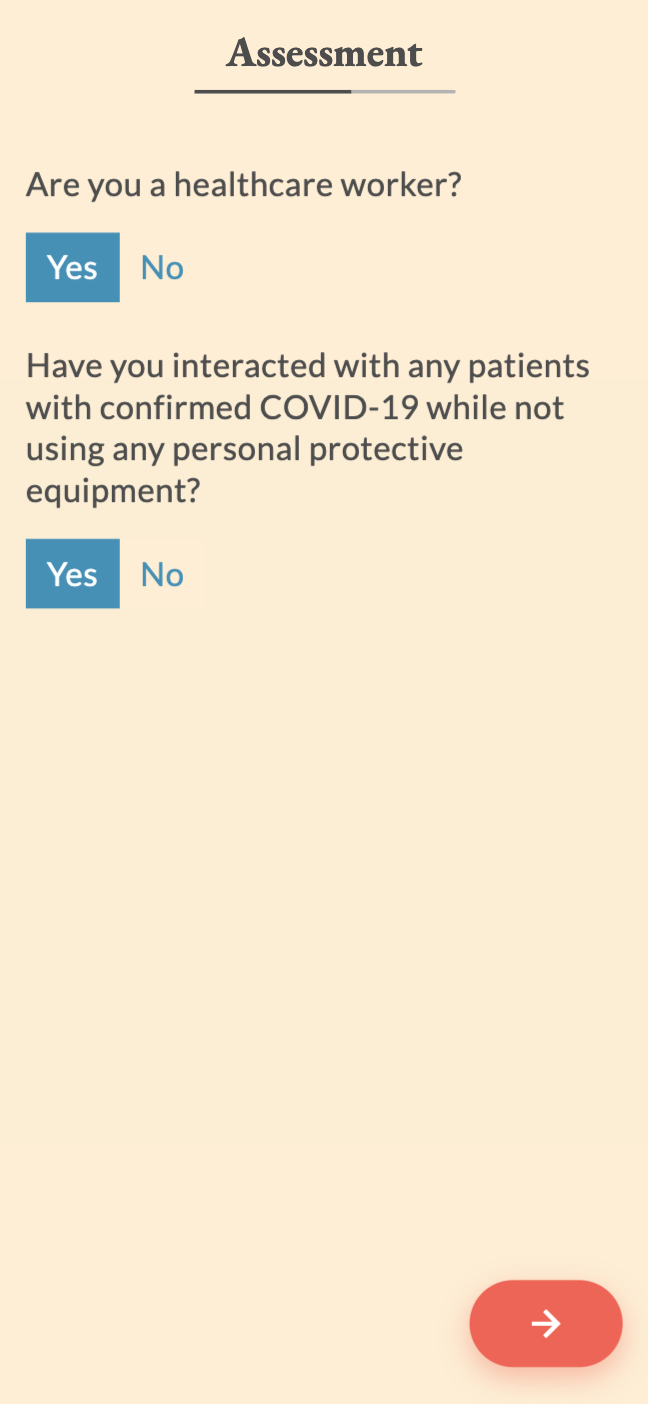}
\endminipage\hfill
\minipage{0.25\textwidth}%
  \includegraphics[width=\linewidth]{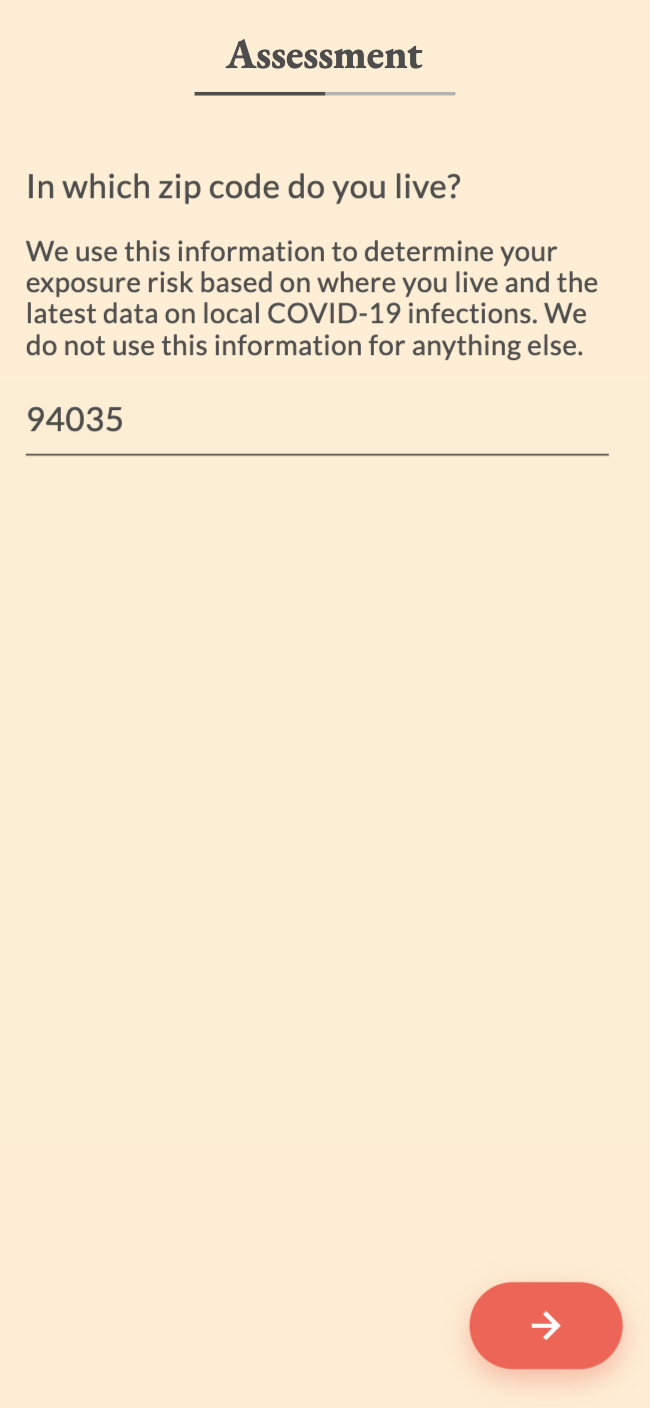}
\endminipage
\caption{Example screenshots from publicly deployed COVID-19 assessment flow from which the data for our work is derived.}
\label{fig:flow}
\end{figure}

In Fig.~\ref{fig:flow} we provide the screenshots from an online telehealth service from which we obtained the COVID assessment data. Tbl.~\ref{fig:covidsim} provides
examples of COVID-19 cases that we generated from this tool.  Tbl.~\ref{tbl:covidFindings} provides the list of findings that are uniquely present in this dataset,
which is included in the model \BaseCOVIDFULL.
\label{apd:simulator}

\begin{table*}
\centering
{\begin{tabular}{l}
\toprule
Moderate COVID-19 risk   \\ 
\hline
\begin{minipage} [t] {\textwidth} 
\begin{itemize}
    \item male,  middle age (41 to 70 yrs),
        nasal congestion, nose discharge,
        been in Japan in the past 14 days
        prior to the onset of symptoms
        prolonged exposure to person with
        confirmed COVID-19 case (but not close contact)
        \item female, middle age (41 to 70 yrs), lightheadedness, dyspnea,  chest pain, dyspnea, exertional, foreign travel history,  hospital personnel, dyspnea at rest 
\end{itemize}
\end{minipage}
\end{tabular}}
{\begin{tabular}{l}
\toprule
Low COVID-19 risk   \\ 
\hline
\begin{minipage} [t] {\textwidth} 
\begin{itemize}
    \item female, middle age (41 to 70 yrs), cough, headache, hospital personnel, cigarette smoking, nose discharge
    \item male, middle age (41 to 70 yrs), lightheadedness, cough , nasal congestion,  sore throat, headache, hospital personnel, nose discharge
\end{itemize}
\end{minipage}
\end{tabular}}
\caption{Examples of cases from COVID assessment flow.}\label{fig:covidsim}
\end{table*}

\begin{table*}
\centering
\small{
{\begin{tabular}{l}
\hline
\begin{tabular}[c]{@{}l@{}}
in China in the past 14 days prior to symptoms onset 
\\ in Europe in the past 14 days prior to symptoms onset
\\ in Iran in the past 14 days prior to symptoms onset
\\ in Japan in the past 14 days prior to symptoms onset
\\ dyspnea at rest
\\ hospital personnel
\\ exposure to COVID- 2019 within the last 14 days \\
\hspace{0.5in}  from symptom onset
\\ healthcare contact with confirmed COVID-19 case \\
\hspace{0.5in} not using any personal protective equipment
\\ household contact with confirmed COVID-19 case
\\  prolonged exposure to person with confirmed COVID-19 case \\
\hspace{0.5in}  (but not close contact)
\\ recent close contact with a person with  \\
\hspace{0.5in}  symptomatic confirmed COVID-19 case
\end{tabular}  
\\ \hline
\end{tabular}}}
\caption{Findings unique to COVID-19 assessment data, in comparison to data simulated from expert system.}
\label{tbl:covidFindings}
\end{table*}

\section{\EXPERTTEST}\label{apd:expert}
In Algorithm~\ref{algo:generator}, we provide more details into the algorithm for clinical case simulation from expert system.  While the simulation starts with a single disease, the algorithm itself returns the differential diagnosis, as opposed to the initial starting disease. Differential diagnosis consists of rank-ordered list of diseases, along with their scores under the expert system's inference algorithm.  
Tbl.~\ref{fig:sim}, provides qualitative examples from the simulation.
\input{sections/examples}


\input{sections/algo}

%% file: sections/qual1.tex
\pagebreak

\begin{table*}
\centering
\resizebox{\textwidth}{!}{\begin{tabular}{lllll}
\toprule
Findings   & Findings in our feature space & Label  & \Base & \BaseCOVID  \\ 
\hline

\begin{tabular}[c]{@{}l@{}}
30 y/o m \\ 2 day HX of runny
nose \\ sore throat \\ hot,
sweaty \\ mild headache \\cough
with clear sputum \\muscle
aches \\no fever or neck
stiffness 
\end{tabular}  &
\begin{tabular}[c]{@{}l@{}}
male \\ young adult (18 to 40 yrs) \\brief (6-48 hours)\\nose discharge  \\ sore throat \\  sweating increase \\headache \\productive cough \\ generalized myalgia \\  no fever \\ no neck stiffness
\end{tabular}  &
\begin{tabular}[c]{@{}l@{}}viral upper respiratory \end{tabular}  &
\begin{tabular}[c]{@{}l@{}}{viral upper respiratory (0.415)}\\ influenza (0.298)  \\ acute sinusitis (0.044)\\ streptococcal pharyngitis (0.028)\\bacterial pneumonia (0.026)\end{tabular} &
\begin{tabular}[c]{@{}l@{}}{viral upper respiratory (0.257)}\\ influenza (0.218)\\ {\bf COVID-19} (0.181)  \\ streptococcal pharyngitis (0.029)\\ opioid withdrawal (0.022)\end{tabular} 
\\ \hline

\begin{tabular}[c]{@{}l@{}}
18 y/o m \\ 3 days severe
headache\\ fever\\
photophobia,\\neck stiffness
\end{tabular}  &
\begin{tabular}[c]{@{}l@{}}
male \\ young adult (18 to 40 yrs) \\ severe headache \\fever \\ photophobia \\ neck stiffness
\end{tabular}  &
\begin{tabular}[c]{@{}l@{}}meningitis \end{tabular}  &
\begin{tabular}[c]{@{}l@{}} meningitis (0.869) \\ West Nile fever (0.102) \\  Rocky Mountain spotted fever (0.009)\\ influenza (0.005) \\ relapsing fever (0.002)\end{tabular} &
\begin{tabular}[c]{@{}l@{}} meningitis (0.771) \\ West Nile fever (0.200) \\  Rocky Mountain spotted fever (0.009)\\ influenza (0.005) \\ epidemic myalgic encephalomyelitis (0.002)\end{tabular}
\\ \hline
\end{tabular}}
\caption{Sample model predictions for examples from \Semigran.  Column 1-2 correspond to set of findings provided in \cite{Semigran} and their transformation to findings understood. Column 3 is the ground truth label. Columns 4-5 are the top 5 diseases predicted by models and the corresponding probabilities. We can see that for the first example with `viral upper respiratory infection', COVID-19 is in the top-5 diseases under \BaseCOVID.}
\label{fig:semigranQual}
\end{table*}

%% file: sections/qual2.tex
\begin{table*}
\centering
\resizebox{\textwidth}{!}{\begin{tabular}{lllll}
\toprule
Common symptoms & Specific to \BaseCOVIDFULL &  \Base & \BaseCOVID  & \BaseCOVIDFULL \\ 
\hline

\begin{tabular}[c]{@{}l@{}}
male\\ young adult (18 to 40 yrs)\\fever\\cough
\end{tabular}  &
\begin{tabular}[c]{@{}l@{}}
 prolonged exposure to \\
 \hspace{.1pt}person with confirmed COVID-19 case \\(but not close contact)
\end{tabular}  &
\begin{tabular}[c]{@{}l@{}}  influenza (.882) \\ bacterial pneumonia (.073) \\  common cold (0.029) \\ acute sinusitis (0.006) \\ asthma (0.006)\end{tabular}    &
\begin{tabular}[c]{@{}l@{}}  influenza (.779) \\ bacterial pneumonia (.131) \\ {\bf COVID-19} (0.038) \\  common cold (0.022)  \\ asthma (0.015)\end{tabular}    &
\begin{tabular}[c]{@{}l@{}} {\bf COVID-19} (0.761) \\ influenza (.106) \\ bacterial pneumonia (0.06) \\  common cold (0.018)  \\ acute sinusitis (0.012)
\end{tabular}
\\ \hline
\begin{tabular}[c]{@{}l@{}}
female\\ young adult (18 to 40 yrs)\\chest pain\\cough\\dyspnea\\nasal congestion
\end{tabular}  &
\begin{tabular}[c]{@{}l@{}}
dyspnea at rest \\ been in Europe in the past \\14 days prior to the onset of symptom
\end{tabular}  &
\begin{tabular}[c]{@{}l@{}}  pulmonary embolism (.372) \\ chronic bronchitis (.213) \\  asthma (0.198) \\ bacterial pneumonia (0.101) \\ influenza (0.051)\end{tabular}    &
\begin{tabular}[c]{@{}l@{}}  pulmonary embolism (.258) \\ asthma (0.216) \\chronic bronchitis (.182) \\   bacterial pneumonia (0.151) \\ influenza (0.060)\end{tabular}    &
\begin{tabular}[c]{@{}l@{}} {\bf COVID-19} (0.492) \\ pulmonary embolism (.088) \\acute bronchitis (.084) \\  asthma (0.076) \\ bacterial pneumonia (0.064)\end{tabular}    
\\ \hline
\begin{tabular}[c]{@{}l@{}}
female\\ middle age (41 to 70 yrs)\\cough\\sore throat\\nasal congestion \\foreign travel history
\end{tabular}  &
\begin{tabular}[c]{@{}l@{}}
 None
\end{tabular}  &
\begin{tabular}[c]{@{}l@{}} common cold (0.843) \\ infectious mononucleosis (0.064) \\  influenza (0.029)\\ measles (0.018) \\ acute sinusitis (0.01)\end{tabular} &
\begin{tabular}[c]{@{}l@{}}  {\bf COVID-19} (0.977) \\common cold (0.015) \\  measles (0.003) \\infectious mononucleosis (0.001) \\  influenza (.0003)\end{tabular} &
\begin{tabular}[c]{@{}l@{}} {\bf COVID-19} (0.770) \\common cold (0.145) \\ infectious mononucleosis (0.032) \\   measles (0.0165) \\influenza (.010)\end{tabular}   
\\ \hline

\begin{tabular}[c]{@{}l@{}}
female\\ middle age (41 to 70 yrs)\\generalized myalgia\\headache\\neck stiffness
\end{tabular}  &
\begin{tabular}[c]{@{}l@{}}
hospital personnel
\end{tabular}  &
\begin{tabular}[c]{@{}l@{}} West Nile fever (0.356) \\  bacterial meningitis (0.305) \\Saint Louis encephalitis (0.249) 
\\  Lyme disease (0.069)\\ Rocky Mountain fever(0.005)\end{tabular} &
\begin{tabular}[c]{@{}l@{}}  Saint Louis encephalitis (0.341) \\  West Nile fever (0.308)\\ bacterial meningitis (0.247) \\ Lyme disease (0.073)\\ Rocky Mountain fever(0.011)\end{tabular}  &
\begin{tabular}[c]{@{}l@{}}  Saint Louis encephalitis (0.398) \\  West Nile fever (0.320)\\ Lyme disease (0.078) \\ {\bf COVID-19} (0.053)\\ bacterial meningitis (0.031) \end{tabular}  
\\ \hline
\end{tabular}
}
\\
\caption{Model predictions (along with scores) for test data from COVID-19 risk assessment. Column 1 corresponds to findings used as input in all three models. Column 2 are the additional findings used by only \BaseCOVIDFULL. }
\label{tbl:covidExamples}
\end{table*}

%% file: sections/examples.tex
\begin{table}
\centering
\small{
{\begin{tabular}{ll}
\toprule
Findings   &  Differential Diagnosis \\ 
\hline
\begin{tabular}[c]{@{}l@{}}
female \\ young adult (18 to 40 yrs) \\  few days (2-7 days) \\  cough \\  productive cough \\ malaise \\   fever \\ nose discharge
\end{tabular}  &
\begin{tabular}[c]{@{}l@{}}bacterial pneumonia (26.9) \\ influenza (23.4) \\acute sinusitis (22.9)\end{tabular} 
\\ \hline
\begin{tabular}[c]{@{}l@{}}
male \\ middle age (41 to 70 yrs) \\  prolonged (1-4 weeks) \\ foot pain \\ heel pain \\ movement pain \\ walking pain \\ limping
\end{tabular}  &
\begin{tabular}[c]{@{}l@{}}plantar fasciitis (38.9) \\ stress fracture (31.5)\end{tabular} 
\\ \hline
\begin{tabular}[c]{@{}l@{}}
female  \\  middle age (41 to 70 yrs)  \\   chronic ($>$ 4 weeks)  \\   urinary urgency  \\   change in bladder habits \\   nocturia  \\   urinary frequency
\end{tabular}  &
\begin{tabular}[c]{@{}l@{}} female  urethritis (32.9) \\ overactive bladder (26.9) \end{tabular} 
\\ \hline
\end{tabular}}}
\caption{Examples of clinical cases using simulation from expert system. Raw scores (shown) converted to probabilities using softmax.}\label{fig:sim}
\end{table}

%% file: sections/algo.tex
\begin{algorithm*}
\begin{algorithmic}
\STATE {\bf Input}: Medical knowledge base of diseases ($\mathcal{D}$) and findings ($\mathcal{F}$). Term-frequency $FREQ(d,f)$ with $d \in \mathcal{D} $ and  $f \in \mathcal{F}$; number of cases T. Expert Inference engine $ExpertInference$ that takes us input a set of findings and provides the differential diagnosis. 
\STATE {\bf Output}: $\{{\bf f_{pos}}^{(t)}, {\bf f_{neg}}^{(t)}, {\bf ddx}^{(t)}\}_{t=1}^{T}$ pairs. {\bf ddx} is differential diagnosis consisting of top $K$ pairs of ($y \in \mathcal{D}$,  $s \ge 0)$  .
\FOR{t = 1 to T} 
  \STATE  $y \sim  Uniform(\mathcal{D}^{*}) $ \COMMENT{Candidate disease label for this case}
  \STATE ${\bf f}^{(t)} \gets \emptyset$
  \STATE $F^{*} = sort(\mathcal{F}, FREQ(y,:))$ 
  \FOR{$f \in \textrm{Demographics}(\mathcal{F})$} 
  	\IF{$f \in F^{*}\textrm{and rand()} > FREQ(f,y)$}
  	 	\STATE $ {\bf f_{pos}}^{(t)} \gets {\bf f_{pos}}^{(t)} \cup  \{f\}$
  	 	\STATE $F^{*} \gets \textrm{RemoveMutex}(F^{*} , {\bf f_{pos}}^{(t)})$ 
     \ENDIF
   \ENDFOR
   \STATE $L = randint(5, |F^{*}|) + |  {\bf f_{pos}}^{(t)}|$  
   \STATE $F^{*} \gets  \mathcal{F} \setminus \textrm{Demographics}(\mathcal{F})$ 
    \WHILE{ $| {\bf f}^{(t)}| \leq L $}
    	\STATE $f^{*} \gets \textrm{GetNext}(F^{*})$ 
        \IF {$FREQ(f,y^{(t)}) \ge .2 $  and $rand() < FREQ(f,y^{(t)})$}
    		\STATE $ {\bf f_{pos}}^{(t)}  \gets {\bf f_{pos}}^{(t)} \cup \{f^{*}\} $
            \STATE $F^{*} \gets F^{*} \setminus \{f^{*}\}$
            \STATE $F^{*} \gets \textrm{RemoveMutex}(F^{*}, {\bf f_{pos}}^{(t)})$
        \ELSIF {$FREQ(f,y^{(t)}) < .2 $  and $ \textrm{rand()} >0.75$}
            \STATE $ {\bf f_{neg}}^{(t)}  \gets {\bf f_{neg}}^{(t)} \cup \{f^{*}\} $
            \STATE $F^{*} \gets F^{*} \setminus \{f^{*}\}$
 		\ENDIF 
  	\ENDWHILE
  	\STATE  ${\bf ddx}^{(t)} \gets \textrm{ExpertInference}({\bf f_{pos}}^{(t)}, {\bf f_{neg}}^{(t)})$ 
\ENDFOR
 \end{algorithmic}
   \caption{Clinical case simulation using expert system}
  \label{algo:generator}
\end{algorithm*}